\documentclass{ametsocV6.1}

\title{Perspectives on AI Architectures and Co-design for Earth System Predictability}

\authors{Maruti K. Mudunuru,\aff{1}\correspondingauthor{M.~K.~Mudunuru, Email:~maruti@pnnl.gov; PNNL-SA-183269; LA-UR-23-22945; LLNL-JRNL-846885} 
James A. Ang,\aff{1} 
Mahantesh Halappanavar,\aff{1} 
Simon D. Hammond,\aff{2} 
Maya B. Gokhale,\aff{3}
James C. Hoe,\aff{4}
Tushar Krishna,\aff{5}
Sarat S. Sreepathi,\aff{6}
Matthew R. Norman,\aff{6}
Ivy B. Peng,\aff{7}
and Philip W. Jones\aff{8}
}

\affiliation{\aff{1}{Pacific Northwest National Laboratory, Richland, WA, USA.}\\
\aff{2}{National Nuclear Security Administration (NNSA), Washington, District of Columbia, USA.}\\
\aff{3}{Lawrence Livermore National Laboratory, Livermore, CA, USA.}\\
\aff{4}{Carnegie Mellon University, Pittsburgh, PA, USA.}\\
\aff{5}{Georgia Institute of Technology, Atlanta, GA, USA.}\\
\aff{6}{Oak Ridge National Laboratory, Oak Ridge, TN, USA.}\\
\aff{7}{KTH Royal Institute of Technology, Sweden.}\\
\aff{8}{Los Alamos National Laboratory, Los Alamos, NM, USA.}
}

\abstract{
Recently, the U.S. Department of Energy (DOE), Office of Science, Biological and Environmental Research (BER), and Advanced Scientific Computing Research (ASCR) programs organized and held the Artificial Intelligence for Earth System Predictability (AI4ESP) workshop series. 
From this workshop, a critical conclusion that the DOE BER and ASCR community came to is the requirement to develop a new paradigm for Earth system predictability focused on enabling artificial intelligence (AI) across the field, lab, modeling, and analysis activities, called ModEx.
The BER's `Model-Experimentation', ModEx, is an iterative approach that enables process models to generate hypotheses.
The developed hypotheses inform field and laboratory efforts to collect measurement and observation data, which are subsequently used to parameterize, drive, and test model (e.g., process-based) predictions.
A total of 17 technical sessions were held in this AI4ESP workshop series.
This paper discusses the topic of the `AI Architectures and Co-design' session and associated outcomes.
The AI Architectures and Co-design session included two invited talks, two plenary discussion panels, and three breakout rooms that covered specific topics, including: (1) DOE HPC Systems, (2) Cloud HPC Systems, and (3) Edge computing and Internet of Things (IoT).
We also provide forward-looking ideas and perspectives on potential research in this co-design area that can be achieved by synergies with the other 16 session topics. These ideas include topics such as: (1) reimagining co-design, (2) data acquisition to distribution, (3) heterogeneous HPC solutions for integration of AI/ML and other data analytics like uncertainty quantification with earth system modeling and simulation, and (4) AI-enabled sensor integration into earth system measurements and observations.
Such perspectives are a distinguishing aspect of this paper.
} 

\begin{document}

\maketitle

%
\statement
This study aims to provide perspectives on AI architectures and co-design approaches for Earth-system predictability.
Such visionary perspectives are essential because AI-enabled model-data integration has shown promise in improving predictions associated with climate change, perturbations, and extreme events.
Our forward-looking ideas guide what is next in co-design to enhance Earth-system models, observations, and theory using state-of-the-art and futuristic computational infrastructure.

%
\section{Introduction}
\label{Sec:Intro}
\noindent
The U.S. Department of Energy (DOE) recently concluded a workshop on Artificial Intelligence for Earth-System Predictability (AI4ESP) \citep{hickmon2022artificial}.
This workshop was hosted by the DOE's Office of Science, Biological and Environmental Research (BER) and Advanced Scientific Computing Research (ASCR) Programs. 
A total of 17 sessions with researchers worldwide participated and discussed how artificial intelligence (AI) could enhance Earth-system predictability across the field, lab, modeling, and analysis activities \citep[Fig-1.3]{hoffman20172016}.
The primary focus of the discussion was on using AI for transforming BER's ``Model-Experimentation'' (ModEx) integration \citep[page-93]{chambers2012research}.

Traditionally, the ModEx paradigm \citep[Section-1]{hoffman20172016} integrates observations, experiments, and measurements performed in the field or laboratory with conceptual/process models in an iterative fashion.
Recent advances in AI have shown promise to accelerate the traditional ModEx efficiently \citep{tsai2021calibration,cromwell2021estimating,mudunuruscalable}.
Such a transformation in the ModEx loop is needed to efficiently and accurately integrate the DOE's observational capabilities and platforms\footnote{Popular BER observational capabilities include Atmospheric Radiation Measurement Climate Research Facility (ARM) \citep{ARM2022} and Environmental Molecular Sciences Laboratory (EMSL) \citep{EMSL2022}}, process models and software infrastructure\footnote{State-of-the-art DOE-funded, open-source, and massively-parallel multi-physics codes include \texttt{PFLOTRAN} \citep{pflotran-web-page}, Advanced Terrestrial Simulator (\texttt{ATS}) \citep{ATS2022}, and Energy Exascale Earth System Model (\texttt{E3SM}) \citep{E3SM2022}}, and computational hardware\footnote{ASCR-funded computational infrastructure and scientific user facilities include Argonne Leadership Computing Facility (ALCF) \citep{ALCF2022}, National Energy Research Scientific Computing Center (NERSC) \citep{NERSC2022}, and Oak Ridge Leadership Computing Facility (OLCF) \citep{OLCF2022}}.
However, achieving this AI-enabled ModEx vision requires significant advancements in co-design and associated AI architectures \citep{germann2013exascale,zhang2019neural,beckman20205g,descour2021workshop,bringmann2021automated}.
This paper provides perspectives on AI architectures and co-design approaches needed to develop AI-enabled ModEx for Earth-system predictability.
These perspectives include co-designing computational and storage infrastructure for automated ML feature engineering and model selection, integration of sensors, process models, and ML methods for efficient data assimilation.
We also provide futuristic system ideas on co-designing frameworks and platforms to enable the BER community to accelerate the application of AI architectures in the ModEx lifecycle.

The outline of our paper is as follows:~Sec.~\ref{Sec:StateOfTheArt} presents the state-of-the-science on AI architectures and co-design that AI4ESP workshop participants discussed.
Section~\ref{Sec:FutureSystemConcepts} provides four different futuristic concepts, and Sec.~\ref{Sec:GrandChallenges} discusses the grand challenges of developing such ideas.
We also discuss near-, middle-, and long-term goals to overcome these grand challenges.
Section~\ref{Sec:VisPerspect} provides perspectives for potential research that will provide synergy with other AI4ESP workshop sessions.
Conclusions are drawn in Sec.~\ref{Sec:Conclusions}.

\section{State-of-the-Science}
\label{Sec:StateOfTheArt}
In this section, we describe the state-of-the-science on AI architectures and co-design.
The focus is the computing resources and DOE user facilities used in capturing and curating data, the development of advanced AI/ML models, and inferences for quantifying and improving earth system modeling and simulation predictability.

\subsection{DOE's High-Performance Computing User Facilities}
\label{Subsec:DOE_HPC}
Over the past few decades, DOE has invested hundreds of millions of dollars in developing high-performance computing (HPC) user facilities \citep{stevens2020ai,vetter2022extreme,heroux2022ecp}.
DOE's investments towards exascale computing include Leadership Computing Facilities (LCFs) at Argonne national laboratory (ALCF) (e.g., Aurora), Oak Ridge National Laboratory (OLCF) (e.g., Frontier), and National Energy Research Scientific Computing Center (NERSC), (e.g., Perlmutter).
Frontier is ranked the fastest supercomputing system on the November 2022 Top 500 list.
The latest generations of DOE's leadership-class computing facilities are based on integrating central processing unit (CPU) and graphics processing unit (GPU) processors into heterogeneous systems.
Concurrently, DOE's Biological and Environmental Research Program has invested substantial resources in state-of-the-art scientific models \citep{E3SM2022,pflotran-web-page,ATS2022} including the flagship Energy Exascale Earth System Model (E3SM) \citep{E3SM2022} that is specifically designed to target efficient utilization of the exascale supercomputers. 
These HPC resources have significantly improved model predictability in various areas, including earth system modeling, subsurface flow and transport models, etc. (e.g., \texttt{E3SM}, \texttt{PFLOTRAN}).
As part of the DOE's Exascale Computing Project, a selected subset of earth science applications \cite{E3SM-MMF,Subsurface-Steefel} firmly focused on model development for the exascale era. 
Furthermore, efforts like the E3SM-MMF sub-project \citep{E3SM-MMF} under ECP had targeted co-design activities, including strong engagements with vendors on early architecture evaluation and algorithm design. 
Experience from such efforts indicates the need for expansion to AI architecture co-design and increased coverage of earth science applications.
Such advancements allowed us to achieve energy-efficient performance on GPUs while leveraging the commercial drivers for GPU-based AI/ML performance. 

With the slowing of Moore's Law, the computing community recognized the increased need for architectural specialization.
Hence, the next generation of HPC systems are likely to incorporate increased heterogeneity  beyond the current hybrid CPU and GPU designs. 
The DOE's efforts in AI for Science \citep{baker2019workshop,stevens2020ai} are exploring capabilities that provide a foundation for the integration of HPC applications (e.g., ALCF's AI testbeds \citep{AItestbed2022}) with data science and AI/ML frameworks.

\subsection{Cloud computing}
\label{Subsec:Cloud_Computing}
Cloud providers\footnote{Popular providers include Amazon Web Services (AWS) \citep{AWS2022}, Google Cloud Platform (GCP) \citep{GCP2022}, and Azure \citep{Azure2022}} have user-friendly tools to run AI/ML workloads.
But there needs to be more compatibility among AI/ML tool capabilities and user interfaces among different providers that make it difficult to achieve interoperability in a federation of clouds \citep{chouhan2020survey,rosa2021computational,saxena2021survey}. 
While specific Earth system model (ESM) data are presently stored on cloud storage systems \citep{xu2019using}, the data stores are associated with a patchwork of individual groups and projects, lacking a federated view. 
Cloud providers can presently accommodate petabytes to exabytes of data for data storage. 
The commercial cloud cost is based on accessing and computing or analyzing the data. 
It can become prohibitively expensive if data transmission in to/out of the Cloud becomes frequent. 
Commercial AI/ML cloud infrastructure and services are predominately motivated by text and image data. 
Cloud providers have demonstrated AI-at-Scale for these applications. 
For example, the most significant AI-based Natural Language Processing (NLP) models approaching 1 trillion parameters have been demonstrated on Selene \citep{chen2019selene} (the 9th fastest supercomputing system on the November 2022 Top 500 list). 
Workflow services exist on the Cloud for specific applications, including many AI/ML methods, and raw materials are available on cloud platforms to create more complex workflows. 
However, ESM workflows that combine external data sources or coordinate with HPC simulations efficiently and accurately currently do not exist.
Computer science expertise is required to create such workflows in a form suitable for domain scientists \citep{chen2017running,bauer2021digital}.

\subsection{Edge Computing}
\label{Subsec:Edge_Computing}
Recently, AI methods for classifying patterns, anomaly detection, unsupervised learning for data compression, inference at the edge, and continuous learning with streaming sensor data have gained considerable traction in the ESM community \citep{beckman20205g,talsmafrost}. 
This advancement was possible because of the rapid forward deployment of AI models on intelligent computing devices such as Raspberry Pi/Shake, Nvidia Jetson Nano, Google Coral Dev Board, and Intel Neural Compute Stick connected to sensors. \citep{catlett2017array,catlett2020measuring,mudunuru2021edgeai}.
The integration of edge computing with smart sensors (e.g., AI@SensorEdge) has many distinct
deployment scenarios, including National Oceanic and Atmospheric Administration (NOAA) and National Aeronautics and Space Administration (NASA) earth-observing satellite imagery with edge processing in space or at dedicated ground stations to control DOE's Atmospheric Radiation Measurement (ARM) or Environmental Molecular Science Laboratory (EMSL) user facility instruments \citep{beckman20205g}.
We can also integrate edge computing with the diverse collection of distributed sensors that collect observations and measurements for the DOE's ARM user facility. 
Adaptive sensors with embedded hardware accelerators are now emerging (e.g., Waggle, PurpleAir) \citep{beckman2016waggle,stavroulas2020field,barkjohn2021development}.
New concepts for distributed applications are also under development, such as geomorphic computing, where weather research and forecasting models are distributed, federated, and able to adapt dynamically to the environment \citep{daepp2022eclipse}.

\section{Future System Concepts}
\label{Sec:FutureSystemConcepts}
In this section, we describe several plausible future systems concepts that participants in the breakout room focus groups discussed in the AI4ESP workshop. 
The focus was on the evolution of DOE’s Leadership Computing Facility (LCF) systems for HPC and AI. 
These large-scale heterogeneous computing systems provide a foundation for advancing AI architectures and co-design using HPC.
Moreover, these future concepts have the potential to provide a radically different approach to future earth system modeling and AI-enabled ModEx.

\subsection{Centralized Large-scale HPC Concept}
\label{Subsec:Centralized_HPC}
The baseline system concept is the future evolution of large-scale HPC and cloud computing systems. 
This next step will extend post-exascale architectures beyond the first generation of DOE's heterogeneous systems integrating CPUs and GPUs. 
As the HPC and Cloud computing communities increasingly rely on hardware specialization to improve performance, co-design approaches will support the development of accelerators \citep{lie2021multi,reuther2021ai,cortes2021sally} for frequently used kernels in scientific modeling and AI/ML methods. 
New specialized accelerators may arise to support additional data science capabilities such as uncertainty quantification, streaming analytics, or graph analysis \citep{halappanavar2021graph,acer2021exagraph}. 
These future large-scale computing systems with extreme heterogeneity must be co-designed to support the increased computational and dataset sizes associated with earth science predictability and scientific machine reasoning \citep{yang201610m,zhang2020optimizing,yu2022characterizing}.

\subsection{Edge sensors with Centralized HPC/Cloud Resources Concept}
\label{Subsec:Edge_HPC}
In the second system concept, environmental data are recorded from a broad collection of point \citep{christensen2017raspberry,winter2021monitoring} and distributed sensors (e.g., fiber optics) \citep{lindsey2019illuminating} spread across the globe. 
These advanced sensors are designed to monitor specific items of interest (e.g., river flow, nutrients, temperature, chemical concentration, light) and to communicate these data back to a centralized location \citep{beckman20205g}. 
At this centralized facility, large HPC or cloud computing environments will process the incoming data streams for integration into online simulations of extreme weather events, climate, hydrology, and their impacts on earth systems.

We could utilize AI/ML capabilities within this system concept at multiple points. 
First, the velocity of sensor data coming into the system will potentially overrun even the most significant data processing centers' capabilities.
Hence, such a volume of data is unlikely to be able to be stored in memory or even temporary storage resources (such as file systems or object stores). 
Advanced AI/ML models could be trained and tailored to summarize or select relevant features from the incoming data streams.
Such an encoding or feature selection process will significantly reduce the amount of data that needs to be kept and integrated into ongoing simulations. 
Another potential is for AI/ML models to identify anomalies or precursors \cite{yuan2019using} from the incoming data streams that might suggest areas of interest for simulations to be focused on – for instance, the start of a hurricane or the high likelihood of significant rain-on-snow events or wildfires.

\begin{figure}
  \centering
    \includegraphics[width = 0.85\textwidth]{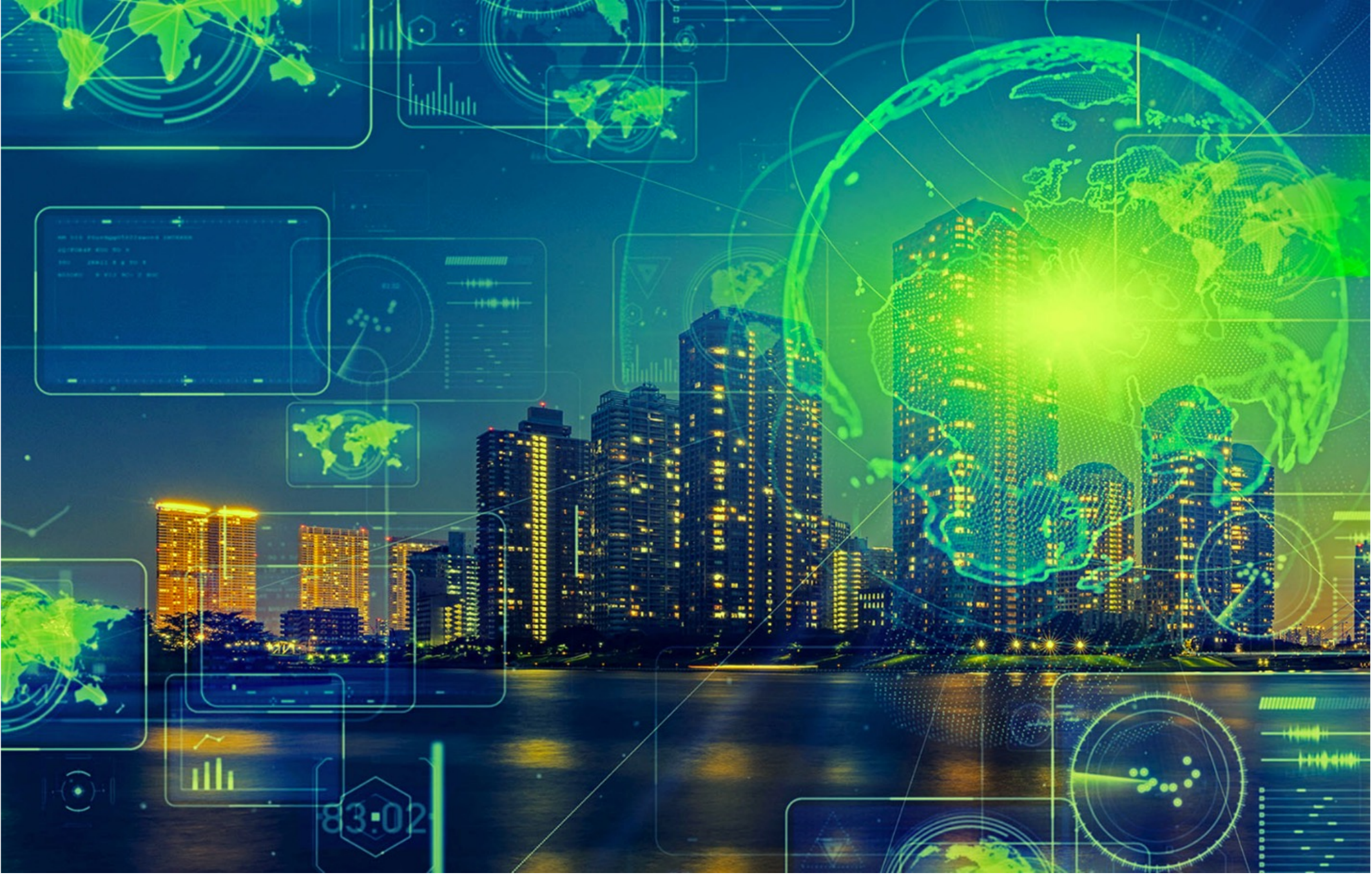}
  \caption{A smart city scenario with a large number of sites for fixed sensor deployments that measure temperature, wind profile, $\mathrm{CO}_2$ concentration, precipitation, etc., plus a variety of mobile devices that can also be used to augment the collection of measurement and observation data intermittently. An urban setting will support advanced wireless communications like 5G and eventually 6G to understand the interactions between cities and climate.
  [Figure developed by Advanced Wireless Communications lab at PNNL]
  \label{Fig:Smart_City}}
\end{figure}

Due to the distributed nature and inhospitable environments (e.g., remote locations, extreme temperatures, or pressures) where sensors may need to be placed or roam, it is unlikely that a reliable data stream will reach the centralized location for all possible inputs. 
One common use case is the intelligent city scenario to study urban science.
Figure~\ref{Fig:Smart_City} is a notional depiction of various deployed sensors, computing, and data storage capabilities \citep{zhu2021sensing}. 
AI/ML models could be used in such an environment to fill measurement gaps and present a more consistent view of observational data to a future simulation run on a large-compute resource.

\subsection{Federated Processing from the Edge to the Data Center Concept}
\label{Subsec:Federated_HPC}
The third potential system design extends the second concept by leveraging much more processing in or near the distributed sensor network. 
We can process the sensor data directly on the sensor itself or in a nearby edge server (e.g., fog computing) with processing elements that may stream a small collection of sensor data into it \citep[Chapter-15]{stevens2020ai}. 
Local processing stations can then send their raw or locally processed data to a centralized HPC and/or cloud resource for inclusion in simulation models and centralized AI/ML models as in the first system concept.

The advantage of this approach is that data down-selection and feature extraction can be performed locally, significantly reducing the volume of data that must be transmitted to a centralized resource. 
Assuming that a sufficiently performant local network among sensors can be established, process model parameters and partial results, perhaps even AI/ML model updates, can be exchanged within a locale, allowing for a genuinely federated design aspect.
Initially, this concept takes advantage of existing gateways and local area networks serving sensors in the field. 
Through co-design collaborations, it is possible to expand that service to include application/sensor-specific processing to filter, analyze, compress, encrypt, and unify multiple sensor streams transmitting measurements through the wireless network.

\subsection{Dynamic and Adaptive Federated Processing Concept}
\label{Subsec:DynamicFed_HPC}
The last system concept builds on the previous three by augmenting feedback and control paths within distributed networks of sensor-local resources \citep{di2021dynamic,charles2021large}. 
Local control offers lower latency decision-making to dynamically control what information is observed, measured, recorded, and relayed by the sensor network \citep{morell2022dynamic}. 
Such a design has powerful implications -- By dynamically controlling sensors online, simulations of the earth’s weather and climate can essentially focus sensor inputs on specific quantities or geographic locations of interest. 
Examples might include where severe weather events are expected or whether climate scientists identify where specific information is needed to help improve the quality of their models. 
This concept expands to multiple HPC and/or Cloud data centers for federated AI/ML modeling. 
AI/ML models can play a crucial part in this system by performing continuous, autonomous online inspection of evolving simulations or recorded data to identify areas of data insufficiency or statistical weakness. 
Furthermore, a dynamic and adaptive system may be able to carefully obtain and select data to improve the quality of its training, reducing the need for vast, potentially intractable datasets to be collected over long periods \citep{catlett2017array}.

\section{Grand Challenges}
\label{Sec:GrandChallenges}
The system concepts that integrate federated processing are beyond the capabilities of affordable technologies today.
It will require significant investment both in foundational technology systems and co-design programs.
Such synergy between climate scientists, mathematicians, AI/ML
experts, computer scientists, and hardware engineers is needed to balance the competing performance, energy, cost, and security challenges associated with AI-enabled ModEx.
The following subsections describe technical challenges that will arise in the areas: (1) programmability and usability, (2) data movement, (3) energy efficiency, and (4) privacy and security of data.

\subsection{Programmability and Usability}
\label{Subsec:ProgUse}
The current and near-term challenge is integrating scientific modeling and simulation applications with AI/ML methods. 
This drives the need to integrate earth system HPC applications written in C/C++ and/or Fortran with AI/ML methods that use Python-based ML frameworks \citep{ott2020fortran}. 
Programming models are under development to support the convergence of applications and workflows onto heterogeneous computing systems. 
Many AI/ML architectures provide hardware support for reduced or mixed precision, and tools will be required to analyze which specific model components can use these capabilities. 
We must create protocols and tools for ESP data-sharing and data federation on the cloud. 
The usability challenge is managing the complexity of mapping converged application workloads to future heterogeneous computing architectures that integrate specialized hardware accelerators with commodity CPU/GPU/TPU processors.

Domain scientists are interested in exploring the capabilities of new heterogeneous advanced architecture computing systems. 
Interfacing with sensors and AI analytics at the Edge will allow domain scientists to extract actionable information needed for improved modeling of disturbances and extreme events. 
This type of co-design is needed for most ESP applications. 
For example, watershed science, hydrology, ecohydrology, climate variability and extremes, aerosols and clouds, and atmospheric modeling are cross-cutting themes where AI\@SensorEdge has the highest impact. 
Co-design approaches that interface with distributed sensor networks will allow us to (1) collect reliable and relevant watershed data under disturbances, (2) monitor land-atmosphere-coastal interactions by embedding intelligence on the Atmospheric Radiation Measurement (ARM) instruments, (3) understand wildfire events and their impact on ecosystems in near-real-time, and (4) assess critical infrastructure impacted by extreme events (e.g., see Human Systems and Dynamics, chapter 9 in AI4ESP report).

Still, there are challenges in understanding how to map AI4ESP workflows to the diverse collection of computing system options. 
Understanding how AI/ML capabilities originally developed for generic commercial workloads may or may not be applicable for Earth System Predictability (ESP) hybrid modeling applications or observation and measurement capabilities is essential. 
From centralized large-scale modeling and training to edge computing inferencing and federated learning, new challenges arise for the composition and distribution of applications, algorithms, and methods. 
This is an important opportunity for the AI4ESP community to develop a new generation of proxy applications and benchmarks for modeling and observation capabilities. 
For example, AI-enabled co-design will enable us to emulate and deploy DOE codes such as PFLOTRAN, ATS, and E3SM at the sensor edge for empowering ARM instruments and EMSL user facilities.
The focus should be facilitating communication and co-design collaborations with hardware designers, system software developers, algorithm developers, and domain scientists.

\subsection{Data Movement}
\label{Subsec:DataMovement}
The expected volume of data associated with a complete, coordinated earth sensor capability will be unprecedented. 
Not only will such a network generate a previously unimaginable quantity and diversity of data, but the computing and network load for processing, transmitting, and subsequent storage of this volume will be orders of magnitude higher than any system available today. 
Data movement costs in terms of energy and latency motivate the interest in the federation and distribution of computing across the AI4ESP scientific ecosystem. 
AI/ML technologies could help reduce such volumes by identifying patterns and anomalies and summarizing sub-volume. 
We will require significant investment in AI/ML approaches to ensure that the modeling capabilities will be compatible and efficient for the types of data being recorded, especially where this may deviate from commercial photo or video capabilities.
Technologies that may assist in energy-efficient data transfers include investment in silicon photonic network capabilities and wide-area 5G- or 6G-like communication networks that enable sensors to communicate over short/medium distances without needing physical wiring \citep{beckman20205g}. 
On the storage side, cloud technologies such as high-performance, large-volume data object stores could likely provide a capability to address increased sparse data storage volumes. 
However, this would pose a significant cost barrier using current commercial cloud pricing. 
We may also use AI/ML to enable innovative compression techniques on earth system data to increase information density without increasing storage costs.
Additionally, DOE HPC centers could incorporate concepts and methods from cloud storage systems into future parallel file and storage systems to slowly move toward such capability.

\subsection{Energy Efficiency}
\label{Subsec:EnergyEfficiency}
Large-scale networks with integrated sensors, federated processing, and wide-area communication networks to handle data transmissions will likely be very expensive in energy consumption. 
While this was a lower-priority focus for exascale computing, data processing and communication remain power-expensive. 
Co-design has the potential to help improve this situation through the use of novel materials, devices, and processing techniques (e.g., neuromorphic-based accelerators to analyze images/video). 
However, significant investment will still be required in foundational technologies if large-scale, power-efficient sensing networks are to be realized. 
Co-design to balance performance and energy efficiency will also address how the modeling, machine learning, uncertainty quantification, and other streaming analytics capabilities are partitioned across the ESM scientific ecosystem.
Such a co-design that integrates DOE’s heterogeneous HPC systems with cloud computing, edge servers, and sensors with IoT devices will transform the ModEx loop.

\subsection{Privacy and Security of Data}
\label{Subsec:PrivacySecurity}
As earth systems modeling becomes increasingly integrated with a distributed network of observations and perhaps federated processing capabilities.
The information's quality, accuracy, and robustness through such a sensor network will become more critical. 
It must also be secured if the information generated from modeling and measurement capabilities is used to support high-consequence national or international scientific policy decisions. 
The implications of potential data tampering or nefarious modification are clear, as a national or international resource for accurate scientific prediction could be severely affected. 
Data privacy concerns are particularly valid in a  data acquisition system where individual human subject images or videos may be captured, or their behavior discerned from the data.
An example includes sensor capabilities that could identify patterns in human systems data (e.g., in citizen science or urban environments). 
Co-design has a potential role in this space -- by including security experts in cyber-physical designs from the outset, secure data transmission and processing can be integrated as a first-level citizen rather than as a last, software-derived additional layer. 
In addition, data privacy may be afforded if local artifacts associated with specific individuals can be aggregated into a larger, federated model with individual patterns obfuscated or redacted into the complete model of the system.

\section{Synergy with other AI4ESP Workshop Sessions}
\label{Sec:VisPerspect}
In this section, we provide visionary perspectives for future ideas and potential research in synergy with other workshop sessions.
Table~\ref{table:Goals} summarizes this synergy with short- ($< 5$ years), medium- (5-year), and long-term (10-year) goals.
The focus is on how AI architectures and co-design approaches are related to the integrative water cycle and associated water cycle extremes.

\begin{table}[htbp]
    \centering
    \caption{This table provides short-, medium-, and long-term goals needed to overcome the grand challenges discussed in Section~\ref{Sec:GrandChallenges}.
    Gradual progress on these specific goals will allow us to advance on the future system concepts needed for improving earth system predictability.}
    \small
    \begin{tabular}{|c|c|c|c|}\hline
        \textbf{Short-term goals} & \textbf{Medium-term goals} & \textbf{Long-term goals} & \textbf{Co-design opportunities} \\ \hline
        Benchmark datasets & Data formats for federated learning & Improve efficiency across ESP domains & Anomaly analysis for extreme events \\ \hline
        Distributed AI/ML workflows & AI/ML for UQ & AI-at-scale demonstration & AI for down- and up-scaling \\ \hline
        AI/ML surrogates & AI/ML + physics simulators & AI for streaming analytics & AI/ML + IoT + Exascale ecosystem \\ \hline
        AI/ML abstractions for edge & AI@SensorEdge & AI-enabled automation & Digital Twin for ESP \\ \hline
    \end{tabular}
    \label{table:Goals}
\end{table}

\textbf{Atmospheric modeling} -- Need for advancing the modeling of subgrid physics across scales and guiding or automating process model calibration. 
This includes  (1) co-design approaches for parameterization and knowledge transfer across scales and (2) AI infrastructure for datasets, software, testing, validation, and training workflows for efficient model calibration.

\textbf{Land modeling} -- AI architectures for efficient transfer of information between land and atmospheric models.
This includes (1) subgrid parameterizations to capture the full complexity within a grid, (2) capturing heterogeneity utilizing LCFs, and (3) addressing observational gaps using advanced AI architectures (e.g., transformers).

\textbf{Hydrology} -- Advanced AI architectures are needed for parameter estimation, down-scaling, and imputation to improve data products. 
Model-data co-design approaches are needed to identify how many and what types of observations are required to reach a desired process model performance without actual measurements being available.
This includes 5G or other high-speed networking or software pipelines that can accelerate the transfer of information between field instrumentation and process models for near real-time sampling decisions.

\textbf{Watershed science} -- Co-design approaches are needed to understand better (1) the quality of collected data, (2) the predictability of a watershed's response (e.g., the evolution of microbial activity) under disturbances and long-term perturbations using process-based models (e.g., \texttt{PFLOTRAN}), (3) when, how, and where to collect data (e.g., wildfires, flooding, drought events), and (4) how to deal with large data volumes.

\textbf{Ecohydrology} -- Advanced AI architectures are needed for developing new data products and benchmark datasets across spatial scales from microbial
and leaf scales to watershed and continental scales.
Novel co-design approaches that build and collect labeled earth science data needed for process models and open-sourcing them to the BER community would facilitate rapid testing of existing AI/ML methods.

\textbf{Aerosols and clouds} -- Co-design approaches that can extract valuable information or identify indicator patterns of forced changes and emergent properties of the actual and simulated climate system are essential.
Future system concepts that can develop databases for indicator
patterns (e.g., nucleation of ice or particles, snow formation) and emergent properties provide a path toward knowledge discovery and reveal missing mechanisms that must be incorporated in process models.

\textbf{Coastal dynamics, oceans, and ice} --  Advanced AI architectures that can improve (1) the standardization and merging of disparate datasets, (2) scale-awareness and dependency in process models (e.g., capturing coastal, ocean, and cryosphere processes across scales and from sparse datasets).

\textbf{Climate variability and extremes} -- Co-design approaches for climate variability, signal identification, and sources of predictability are essential.
These include AI architectures to detect signatures and features corresponding to tropical cyclones, fronts, atmospheric rivers, hailstones, tornadoes, and ice storms.

\textbf{Human systems and dynamics} -- Co-design approaches that can provide a better understanding of human and earth systems. 
For example, advancements in AI architectures are needed to gain better insights into urban prediction and long-term urban policy due to extreme events.

\section{Conclusions}
\label{Sec:Conclusions}
In this perspective paper, we have described the need for co-design approaches for efficient and accurate integration of process models and observations for improved earth system predictability.
Current state-of-science and HPC facilities provide a starting point to address the grand challenges of the `Model-Experimentation' loop.
Future system concepts that connect the edge sensors to intelligent computing devices and, subsequently, the process models that reside in fog/cloud/exascale infrastructure are needed to transform the ModEx lifecycle.
Our near-term to long-term goals allows us to develop AI architectures and co-design approaches using future system concepts.
Community integration and effort between domain and computational experts allow us to transform how we model the integrative and associated water cycle extremes.

\section*{Nomenclature}
\begin{itemize}
  \item AI4ESP:~The Artificial Intelligence for Earth System Predictability
  \item AI:~Artificial Intelligence
  \item ALCF:~Argonne Leadership Computing Facility
  \item ARM:~Atmospheric Radiation Measurement Climate Research Facility
  \item ASCR:~Advanced Scientific Computing Research
  \item \texttt{ATS}:~Advanced Terrestrial Simulator
  \item AWS:~Amazon Web Services
  \item \texttt{E3SM}:~Energy Exascale Earth System Model
  \item ESM:~Earth System Model
  \item ESP:~Earth System Predictability
  \item EMSL:~Environmental Molecular Sciences Laboratory
  \item BER:~Biological and Environmental Research
  \item CPU:~Central Processing Unit
  \item DOE:~Department of Energy
  \item GCP:~Google Cloud Platform
  \item GPU:~Graphics Processing Unit
  \item HPC:~High-Performance Computing
  \item IoT:~Internet of Things
  \item LCF:~Leadership Computing Facility
  \item ModEx:~Model-Experimentation
  \item ML:~Machine Learning
  \item NASA:~National Aeronautics and Space Administration
  \item NERSC:~National Energy Research Scientific Computing Center
  \item NLP:~Natural Language Processing
  \item NOAA:~National Oceanic and Atmospheric Administration
  \item OLCF:~Oak Ridge Leadership Computing Facility
  \item TPU:~Tensor Processing Unit
  \item UQ:~Uncertainty Quantification
\end{itemize}

\clearpage
\acknowledgments
The authors acknowledge all the efforts made as part of the Artificial Intelligence for Earth System Predictability (AI4ESP) workshop.
MKM acknowledges the support from the Environmental Molecular Sciences Laboratory, a DOE Office of Science User Facility sponsored by the Biological and Environmental Research program under Contract No. DE-AC05-76RL01830.
PJ's research was supported as part of the Energy Exascale Earth System Model (E3SM) project, funded by the U.S. Department of Energy, Office of Science, and Office of Biological and Environmental Research.
SS and MN research was supported by the Exascale Computing Project (17-SC-20-SC), a collaborative effort of the U.S. Department of Energy Office of Science and the National Nuclear Security Administration.
MBG's research was performed under the auspices of the U.S. Department of Energy by Lawrence Livermore National Laboratory under Contract DE-AC52-07NA27344.
MKM, JAA, and MH acknowledge the contributions of Johnathan Cree and Elena Peterson at PNNL's Advanced Wireless Communications Lab, who developed the figure in this paper.
This manuscript has been authored by Pacific Northwest National Laboratory (PNNL), operated by Battelle Memorial Institute for the U.S. Department of Energy under Contract No. DE-AC05-76RL01830.
This manuscript has been authored by Oak Ridge National Laboratory, operated by UT-Battelle, LLC under Contract No. DE-AC05-00OR22725 with the U.S. Department of Energy.
The US government retains and the publisher, by accepting the article for publication, acknowledges that the US government retains a nonexclusive, paid-up, irrevocable, worldwide license to publish or reproduce the published form of this manuscript or allow others to do so, for US government purposes. 
DOE will provide public access to these results of federally sponsored research in accordance with the DOE Public Access Plan.

%
\datastatement
There is no data developed in this paper

\bibliographystyle{ametsocV6}
\bibliography{references}

\begin{thebibliography}{61}
\providecommand{\natexlab}[1]{#1}
\providecommand{\url}[1]{\texttt{#1}}
\renewcommand{\UrlFont}{\rmfamily}
\providecommand{\urlprefix}{URL }
\expandafter\ifx\csname urlstyle\endcsname\relax
  \providecommand{\doi}[1]{https://doi.org/\discretionary{}{}{}#1}\else
  \providecommand{\doi}{https://doi.org/\discretionary{}{}{}\begingroup
  \urlstyle{rm}\Url}\fi
\providecommand{\eprint}[2][]{\url{#2}}

\bibitem[{Acer et~al.(2021)}]{acer2021exagraph}
Acer, S., and Coauthors, 2021: Exagraph: Graph and combinatorial methods for
  enabling exascale applications. \textit{The International Journal of High
  Performance Computing Applications}, \textbf{35~(6)}, 553--571.

\bibitem[{ALCF(2022)}]{ALCF2022}
ALCF, 2022: {ALCF -- Argonne Leadership Computing Facility}.
  \urlprefix\url{https://www.alcf.anl.gov/}, accessed on:~2022-11-10.

\bibitem[{ARM(2022)}]{ARM2022}
ARM, 2022: {ARM -- Atmospheric Radiation Measurement Climate Research
  Facility}. \urlprefix\url{https://www.arm.gov/}, accessed on:~2022-11-10.

\bibitem[{ATS(2022)}]{ATS2022}
ATS, 2022: {ATS -- The Advanced Terrestrial Simulator}.
  \urlprefix\url{https://github.com/amanzi/ats}, accessed on:~2022-11-10.

\bibitem[{AWS(2022)}]{AWS2022}
AWS, 2022: {AWS -- Amazon Web Services}.
  \urlprefix\url{https://aws.amazon.com/about-aws/}, accessed on:~2022-11-10.

\bibitem[{Azure(2022)}]{Azure2022}
Azure, 2022: {Microsoft Azure:~Cloud Computing Services}.
  \urlprefix\url{https://azure.microsoft.com/en-us/}, accessed on:~2022-11-10.

\bibitem[{Baker et~al.(2019)}]{baker2019workshop}
Baker, N., and Coauthors, 2019: Workshop report on basic research needs for
  scientific machine learning: Core technologies for artificial intelligence.
  Tech. rep., USDOE Office of Science (SC), Washington, DC (United States).
  \doi{10.2172/1478744}.

\bibitem[{Barkjohn et~al.(2021)Barkjohn, Gantt,, and
  Clements}]{barkjohn2021development}
Barkjohn, K.~K., B.~Gantt, and A.~L. Clements, 2021: Development and
  application of a united states-wide correction for pm 2.5 data collected with
  the purpleair sensor. \textit{Atmospheric Measurement Techniques},
  \textbf{14~(6)}, 4617--4637.

\bibitem[{Bauer et~al.(2021)Bauer, Dueben, Hoefler, Quintino, Schulthess,, and
  Wedi}]{bauer2021digital}
Bauer, P., P.~D. Dueben, T.~Hoefler, T.~Quintino, T.~C. Schulthess, and N.~P.
  Wedi, 2021: The digital revolution of earth-system science. \textit{Nature
  Computational Science}, \textbf{1~(2)}, 104--113.

\bibitem[{Beckman et~al.(2016)Beckman, Sankaran, Catlett, Ferrier, Jacob,, and
  Papka}]{beckman2016waggle}
Beckman, P., R.~Sankaran, C.~Catlett, N.~Ferrier, R.~Jacob, and M.~Papka, 2016:
  Waggle: An open sensor platform for edge computing. \textit{2016 IEEE
  SENSORS}, IEEE, 1--3.

\bibitem[{Beckman et~al.(2020)}]{beckman20205g}
Beckman, P., and Coauthors, 2020: 5g enabled energy innovation: Advanced
  wireless networks for science (workshop report). Tech. rep., Argonne National
  Lab.(ANL), Argonne, IL (United States); Northwestern Univ~….
  \doi{10.2172/1606538}.

\bibitem[{Bringmann et~al.(2021)}]{bringmann2021automated}
Bringmann, O., and Coauthors, 2021: Automated hw/sw co-design for edge ai:
  State, challenges and steps ahead: Special session paper. \textit{2021
  International Conference on Hardware/Software Codesign and System Synthesis
  (CODES+ ISSS)}, IEEE, 11--20.

\bibitem[{Catlett et~al.(2020)Catlett, Beckman, Ferrier, Nusbaum, Papka,
  Berman,, and Sankaran}]{catlett2020measuring}
Catlett, C., P.~Beckman, N.~Ferrier, H.~Nusbaum, M.~E. Papka, M.~G. Berman, and
  R.~Sankaran, 2020: Measuring cities with software-defined sensors.
  \textit{Journal of Social Computing}, \textbf{1~(1)}, 14--27.

\bibitem[{Catlett et~al.(2017)Catlett, Beckman, Sankaran,, and
  Galvin}]{catlett2017array}
Catlett, C.~E., P.~H. Beckman, R.~Sankaran, and K.~K. Galvin, 2017: Array of
  things: a scientific research instrument in the public way: platform design
  and early lessons learned. \textit{Proceedings of the 2nd international
  workshop on science of smart city operations and platforms engineering},
  26--33.

\bibitem[{Chambers et~al.(2012)Chambers, Fisher, Hall, Norby, Wofsy,, and
  Stover}]{chambers2012research}
Chambers, J., R.~Fisher, J.~Hall, R.~J. Norby, S.~C. Wofsy, and D.~Stover,
  2012: Research priorities for tropical ecosystems under climate change
  workshop report, june 4-5, 2012. Tech. rep., USDOE Office of Science (SC),
  Washington, DC (United States). Biological and~….
  \urlprefix\url{https://ess.science.energy.gov/wp-content/uploads/2020/12/NGEE-Tropics3webHR.pdf}.

\bibitem[{Charles et~al.(2021)Charles, Garrett, Huo, Shmulyian,, and
  Smith}]{charles2021large}
Charles, Z., Z.~Garrett, Z.~Huo, S.~Shmulyian, and V.~Smith, 2021: On
  large-cohort training for federated learning. \textit{Advances in neural
  information processing systems}, \textbf{34}, 20\,461--20\,475.

\bibitem[{Chen et~al.(2019)Chen, Cofer, Zhou,, and
  Troyanskaya}]{chen2019selene}
Chen, K.~M., E.~M. Cofer, J.~Zhou, and O.~G. Troyanskaya, 2019: Selene: a
  pytorch-based deep learning library for sequence data. \textit{Nature
  methods}, \textbf{16~(4)}, 315--318.

\bibitem[{Chen et~al.(2017)Chen, Huang, Jiao, Flanner, Raeker,, and
  Palen}]{chen2017running}
Chen, X., X.~Huang, C.~Jiao, M.~G. Flanner, T.~Raeker, and B.~Palen, 2017:
  Running climate model on a commercial cloud computing environment: A case
  study using community earth system model (cesm) on amazon aws.
  \textit{Computers \& Geosciences}, \textbf{98}, 21--25.

\bibitem[{Chouhan et~al.(2020)Chouhan, Bansal, Lauhny,, and
  Chaudhary}]{chouhan2020survey}
Chouhan, L., P.~Bansal, B.~Lauhny, and Y.~Chaudhary, 2020: A survey on cloud
  federation architecture and challenges. \textit{Social Networking and
  Computational Intelligence}, Springer, 51--65.

\bibitem[{Christensen and Blanco~Chia(2017)Christensen, and
  Blanco~Chia}]{christensen2017raspberry}
Christensen, B., and J.~Blanco~Chia, 2017: Raspberry shake-a world-wide citizen
  seismograph network. \textit{AGU Fall Meeting Abstracts}, Vol. 2017,
  S11A--0560.

\bibitem[{Cort{\'e}s et~al.(2021)Cort{\'e}s, Moya,, and
  Valero}]{cortes2021sally}
Cort{\'e}s, U., U.~Moya, and M.~Valero, 2021: When sally met harry or when ai
  met hpc. \textit{Supercomputing Frontiers and Innovations}, \textbf{8~(1)},
  4--7.

\bibitem[{Cromwell et~al.(2021)Cromwell, Shuai, Jiang, Coon, Painter, Moulton,
  Lin,, and Chen}]{cromwell2021estimating}
Cromwell, E., P.~Shuai, P.~Jiang, E.~T. Coon, S.~L. Painter, J.~D. Moulton,
  Y.~Lin, and X.~Chen, 2021: Estimating watershed subsurface permeability from
  stream discharge data using deep neural networks. \textit{Frontiers in Earth
  Science}, \textbf{9}, 613\,011.

\bibitem[{Daepp et~al.(2022)}]{daepp2022eclipse}
Daepp, M.~I., and Coauthors, 2022: Eclipse: An end-to-end platform for
  low-cost, hyperlocal environmental sensing in cities. \textit{2022 21st
  ACM/IEEE International Conference on Information Processing in Sensor
  Networks (IPSN)}, IEEE, 28--40.

\bibitem[{Descour et~al.(2021)Descour, Tsao, Stracuzzi, Wakeland, Schultz,
  Smith,, and Weeks}]{descour2021workshop}
Descour, M., J.~Tsao, D.~Stracuzzi, A.~Wakeland, D.~Schultz, W.~Smith, and
  J.~Weeks, 2021: Workshop report: Ai-enhanced co-design for next-generation
  microelectronics: Innovating innovation. Tech. rep., Sandia National
  Lab.(SNL-NM), Albuquerque, NM (United States). \doi{10.2172/1845383}.

\bibitem[{Di~Lorenzo et~al.(2021)Di~Lorenzo, Battiloro, Merluzzi,, and
  Barbarossa}]{di2021dynamic}
Di~Lorenzo, P., C.~Battiloro, M.~Merluzzi, and S.~Barbarossa, 2021: Dynamic
  resource optimization for adaptive federated learning at the wireless network
  edge. \textit{ICASSP 2021-2021 IEEE International Conference on Acoustics,
  Speech and Signal Processing (ICASSP)}, IEEE, 4910--4914.

\bibitem[{E3SM(2022)}]{E3SM2022}
E3SM, 2022: {E3SM -- The Energy Exascale Earth System Model}.
  \urlprefix\url{https://e3sm.org/}, accessed on:~2022-11-10.

\bibitem[{E3SM-MMF(2022)}]{E3SM-MMF}
E3SM-MMF, 2022: {E3SM Multiscale Modeling Framework}.
  \urlprefix\url{https://www.exascaleproject.org/research-project/e3sm-mmf/},
  accessed on:~2022-11-10.

\bibitem[{EMSL(2022)}]{EMSL2022}
EMSL, 2022: {EMSL -- The Environmental Molecular Sciences Laboratory}.
  \urlprefix\url{https://www.emsl.pnnl.gov/}, accessed on:~2022-11-10.

\bibitem[{GCP(2022)}]{GCP2022}
GCP, 2022: {GCP -- Google Cloud Platform}.
  \urlprefix\url{https://cloud.google.com/}, accessed on:~2022-11-10.

\bibitem[{Germann et~al.(2013)Germann, McPherson, Belak,, and
  Richards}]{germann2013exascale}
Germann, T.~C., A.~L. McPherson, J.~F. Belak, and D.~F. Richards, 2013:
  Exascale co-design center for materials in extreme environments.
  \doi{10.2172/1116965}.

\bibitem[{Halappanavar et~al.(2021)Halappanavar, Minutoli,, and
  Ghosh}]{halappanavar2021graph}
Halappanavar, M., M.~Minutoli, and S.~Ghosh, 2021: Graph analytics in the
  exascale era. \textit{Proceedings of the 18th ACM International Conference on
  Computing Frontiers}, 209--209.

\bibitem[{Heroux et~al.(2022)}]{heroux2022ecp}
Heroux, M.~A., and Coauthors, 2022: Ecp software technology capability
  assessment report. Tech. rep., Oak Ridge National Lab.(ORNL), Oak Ridge, TN
  (United States). \doi{10.2172/1760096}.

\bibitem[{Hickmon et~al.(2022)Hickmon, Varadharajan, Hoffman, Collis,, and
  Wainwright}]{hickmon2022artificial}
Hickmon, N.~L., C.~Varadharajan, F.~M. Hoffman, S.~Collis, and H.~M.
  Wainwright, 2022: Artificial intelligence for earth system predictability
  (ai4esp) workshop report. Tech. rep., Argonne National Lab.(ANL), Argonne, IL
  (United States). \doi{10.2172/1888810}.

\bibitem[{Hoffman et~al.(2017)}]{hoffman20172016}
Hoffman, F.~M., and Coauthors, 2017: 2016 international land model benchmarking
  (ilamb) workshop report. Tech. rep., USDOE Office of Science, Washington, DC
  (United States). \doi{10.2172/1330803}.

\bibitem[{Lichtner et~al.(2020)}]{pflotran-web-page}
Lichtner, P.~C., and Coauthors, 2020: {PFLOTRAN} {W}eb page.
  Http://www.pflotran.org.

\bibitem[{Lie(2021)}]{lie2021multi}
Lie, S., 2021: Multi-million core, multi-wafer ai cluster. \textit{2021 IEEE
  Hot Chips 33 Symposium (HCS)}, IEEE Computer Society, 1--41.

\bibitem[{Lindsey et~al.(2019)Lindsey, Dawe,, and
  Ajo-Franklin}]{lindsey2019illuminating}
Lindsey, N.~J., T.~C. Dawe, and J.~B. Ajo-Franklin, 2019: Illuminating seafloor
  faults and ocean dynamics with dark fiber distributed acoustic sensing.
  \textit{Science}, \textbf{366~(6469)}, 1103--1107.

\bibitem[{Morell and Alba(2022)Morell, and Alba}]{morell2022dynamic}
Morell, J.~{\'A}., and E.~Alba, 2022: Dynamic and adaptive fault-tolerant
  asynchronous federated learning using volunteer edge devices. \textit{Future
  Generation Computer Systems}, \textbf{133}, 53--67.

\bibitem[{Mudunuru et~al.(2022)Mudunuru, Son, Jiang, Hammond,, and
  Chen}]{mudunuruscalable}
Mudunuru, M.~K., K.~Son, P.~Jiang, G.~Hammond, and X.~Chen, 2022: Scalable deep
  learning for watershed model calibration. \textit{Frontiers in Earth
  Science}, 2206.

\bibitem[{Mudunuru et~al.(2021)}]{mudunuru2021edgeai}
Mudunuru, M.~K., and Coauthors, 2021: {EdgeAI}: How to use ai to collect
  reliable and relevant watershed data. Tech. rep., Artificial Intelligence for
  Earth System Predictability (AI4ESP~….
  \doi{https://doi.org/10.2172/1769700}.

\bibitem[{NERSC(2022)}]{NERSC2022}
NERSC, 2022: {NERSC -- National Energy Research Scientific Computing Center}.
  \urlprefix\url{https://www.nersc.gov/}, accessed on:~2022-11-10.

\bibitem[{OLCF(2022)}]{OLCF2022}
OLCF, 2022: {OLCF -- Oak Ridge Leadership Computing Facility}.
  \urlprefix\url{https://www.olcf.ornl.gov/}, accessed on:~2022-11-10.

\bibitem[{Ott et~al.(2020)Ott, Pritchard, Best, Linstead, Curcic,, and
  Baldi}]{ott2020fortran}
Ott, J., M.~Pritchard, N.~Best, E.~Linstead, M.~Curcic, and P.~Baldi, 2020: A
  fortran-keras deep learning bridge for scientific computing.
  \textit{Scientific Programming}, \textbf{2020}.

\bibitem[{Reuther et~al.(2021)Reuther, Michaleas, Jones, Gadepally, Samsi,, and
  Kepner}]{reuther2021ai}
Reuther, A., P.~Michaleas, M.~Jones, V.~Gadepally, S.~Samsi, and J.~Kepner,
  2021: Ai accelerator survey and trends. \textit{2021 IEEE High Performance
  Extreme Computing Conference (HPEC)}, IEEE, 1--9.

\bibitem[{Rosa et~al.(2021)Rosa, Ralha, Holanda,, and
  Araujo}]{rosa2021computational}
Rosa, M.~J., C.~G. Ralha, M.~Holanda, and A.~P. Araujo, 2021: Computational
  resource and cost prediction service for scientific workflows in federated
  clouds. \textit{Future Generation Computer Systems}, \textbf{125}, 844--858.

\bibitem[{Saxena et~al.(2021)Saxena, Gupta,, and Singh}]{saxena2021survey}
Saxena, D., R.~Gupta, and A.~K. Singh, 2021: A survey and comparative study on
  multi-cloud architectures: emerging issues and challenges for cloud
  federation. \textit{arXiv preprint arXiv:2108.12831}.

\bibitem[{Stavroulas et~al.(2020)}]{stavroulas2020field}
Stavroulas, I., and Coauthors, 2020: Field evaluation of low-cost pm sensors
  (purple air pa-ii) under variable urban air quality conditions, in greece.
  \textit{Atmosphere}, \textbf{11~(9)}, 926.

\bibitem[{Stevens et~al.(2020)Stevens, Taylor, Nichols, Maccabe, Yelick,, and
  Brown}]{stevens2020ai}
Stevens, R., V.~Taylor, J.~Nichols, A.~B. Maccabe, K.~Yelick, and D.~Brown,
  2020: Ai for science: Report on the department of energy (doe) town halls on
  artificial intelligence (ai) for science. Tech. rep., Argonne National
  Lab.(ANL), Argonne, IL (United States). \doi{10.2172/1604756}.

\bibitem[{Subsurface-ECP(2022)}]{Subsurface-Steefel}
Subsurface-ECP, 2022: {Subsurface:~An Exascale Subsurface Simulator of Coupled
  Flow, Transport, Reactions, and Mechanics}.
  \urlprefix\url{https://www.exascaleproject.org/research-project/subsurface/},
  accessed on:~2022-11-10.

\bibitem[{Talsma et~al.(2022)Talsma, Solander, Mudunuru, Crawford,, and
  Powell}]{talsmafrost}
Talsma, C., K.~C. Solander, M.~K. Mudunuru, B.~Crawford, and M.~Powell, 2022:
  Frost prediction using machine learning and deep neural network models for
  use on {IoT} sensors.

\bibitem[{Testbed(2022)}]{AItestbed2022}
Testbed, A.~A., 2022: {ALCF AI Testbed, a next generation of AI-accelerator
  machines}. \urlprefix\url{https://www.alcf.anl.gov/alcf-ai-testbed}, accessed
  on:~2022-11-10.

\bibitem[{Tsai et~al.(2021)Tsai, Feng, Pan, Beck, Lawson, Yang, Liu,, and
  Shen}]{tsai2021calibration}
Tsai, W.-P., D.~Feng, M.~Pan, H.~Beck, K.~Lawson, Y.~Yang, J.~Liu, and C.~Shen,
  2021: From calibration to parameter learning:~harnessing the scaling effects
  of big data in geoscientific modeling. \textit{Nature communications},
  \textbf{12}, 1--13.

\bibitem[{Vetter et~al.(2022)}]{vetter2022extreme}
Vetter, J.~S., and Coauthors, 2022: Extreme heterogeneity 2018-productive
  computational science in the era of extreme heterogeneity: Report for doe
  ascr workshop on extreme heterogeneity. \doi{10.2172/1473756}.

\bibitem[{Winter et~al.(2021)Winter, Lombardi, Diaz-Moreno,, and
  Bainbridge}]{winter2021monitoring}
Winter, K., D.~Lombardi, A.~Diaz-Moreno, and R.~Bainbridge, 2021: Monitoring
  icequakes in east antarctica with the raspberry shake. \textit{Seismological
  Research Letters}, \textbf{92~(5)}, 2736--2747.

\bibitem[{Xu et~al.(2019)Xu, Wei, Dennis,, and Paul}]{xu2019using}
Xu, H., W.~Wei, J.~Dennis, and K.~Paul, 2019: Using cloud-friendly data format
  in earth system models. \textit{AGU Fall Meeting Abstracts}, Vol. 2019,
  IN13C--0728.

\bibitem[{Yang et~al.(2016)}]{yang201610m}
Yang, C., and Coauthors, 2016: 10m-core scalable fully-implicit solver for
  nonhydrostatic atmospheric dynamics. \textit{SC'16: Proceedings of the
  International Conference for High Performance Computing, Networking, Storage
  and Analysis}, IEEE, 57--68.

\bibitem[{Yu et~al.(2022)}]{yu2022characterizing}
Yu, Y., and Coauthors, 2022: Characterizing uncertainties of earth system
  modeling with heterogeneous many-core architecture computing.
  \textit{Geoscientific Model Development Discussions}, 1--23.

\bibitem[{Yuan et~al.(2019)}]{yuan2019using}
Yuan, B., and Coauthors, 2019: Using machine learning to discern eruption in
  noisy environments: A case study using co2-driven cold-water geyser in
  chimay{\'o}, new mexico. \textit{Seismological Research Letters},
  \textbf{90~(2A)}, 591--603.

\bibitem[{Zhang et~al.(2020)}]{zhang2020optimizing}
Zhang, S., and Coauthors, 2020: Optimizing high-resolution community earth
  system model on a heterogeneous many-core supercomputing platform.
  \textit{Geoscientific Model Development}, \textbf{13~(10)}, 4809--4829.

\bibitem[{Zhang et~al.(2019)Zhang, Jiang, Shi,, and Hu}]{zhang2019neural}
Zhang, X., W.~Jiang, Y.~Shi, and J.~Hu, 2019: When neural architecture search
  meets hardware implementation: from hardware awareness to co-design.
  \textit{2019 IEEE Computer Society Annual Symposium on VLSI (ISVLSI)}, IEEE,
  25--30.

\bibitem[{Zhu et~al.(2021)Zhu, Shen,, and Martin}]{zhu2021sensing}
Zhu, T., J.~Shen, and E.~R. Martin, 2021: Sensing earth and environment
  dynamics by telecommunication fiber-optic sensors: An urban experiment in
  pennsylvania, usa. \textit{Solid Earth}, \textbf{12~(1)}, 219--235.

\end{thebibliography}
\end{document}